\pdfoutput=1

\documentclass[11pt]{article}

\usepackage{acl}

\usepackage{times}
\usepackage{latexsym}
\usepackage{graphicx}

\usepackage[T1]{fontenc}

\usepackage[utf8]{inputenc}

\usepackage{microtype}

\title{'Tis but Thy Name: Semantic Question Answering Evaluation\\with 11M Names for 1M Entities}

\author{Albert Huang \\
  The Nueva School \\
  \texttt{albhuan@nuevaschool.org} \\
  }

\begin{document}
\maketitle
\begin{abstract}
Classic lexical-matching-based QA metrics are slowly being phased out because they punish succinct or informative outputs just because those answers were not provided as ground truth. Recently proposed neural metrics can evaluate semantic similarity but were trained on small textual similarity datasets grafted from foreign domains. We introduce the Wiki Entity Similarity (WES) dataset, an 11M example, domain targeted, semantic entity similarity dataset that is generated from link texts in Wikipedia. WES is tailored to QA evaluation: the examples are entities and phrases and grouped into semantic clusters to simulate multiple ground-truth labels. Human annotators consistently agree with WES labels, and a basic cross encoder metric is better than four classic metrics at predicting human judgments of correctness. 
\end{abstract}

\section{Introduction}

Information retrieval tools have already revolutionized how we interact with knowledge: what would have taken a half-hour library trip can be learned in minutes through the internet. Question answering (QA) models can take this one step further because they can respond to queries with summative insights instead of just listing relevant documents. Because they dictate the optimization goal of QA research, QA evaluation metrics are a key in directing the future of QA model development.

Classic question-answering evaluation metrics only consider the token overlap between a model's output and human-annotated ground-truth answers. However, these evaluation metrics fail to consider the plurality of possible correct answers for every question. For instance, a “when” question can be answered using either a year or an event name (i.e. “During Super Bowl XXXI. ”) Furthermore, correct answers can have wildly different lexical signatures if they contain different words, phrasing, or level of detail. Existing QA metrics also cannot evaluate against multiple ground-truth annotations simultaneously, leading to underutilization of dataset resources. Therefore, metrics that rely on the tokens in QA model outputs unfairly punish non-conforming models, including those that provide creatively succinct or informative answers. These metrics force models to imitate the threadbare, minimal ground-truth answers that crowd workers tend to write, exacerbating the problem. Current QA metrics are limiting the usefulness of QA models. With the rise of generative QA, in which models have more freedom to use different words or phrasing than extractive QA, the need for semantic answer evaluation has only grown.

Previous semantic techniques used latent word or phrase embeddings to calculate the similarity between answers. More recently, end-to-end neural text similarity models have been applied to QA evaluation. However, they are trained on small datasets grafted from other tasks. In particular, QA models often answer questions about specific entities or phrases; there are currently no entity or short phrase semantic similarity datasets.

\begin{figure}
  \includegraphics[width=0.48\textwidth]{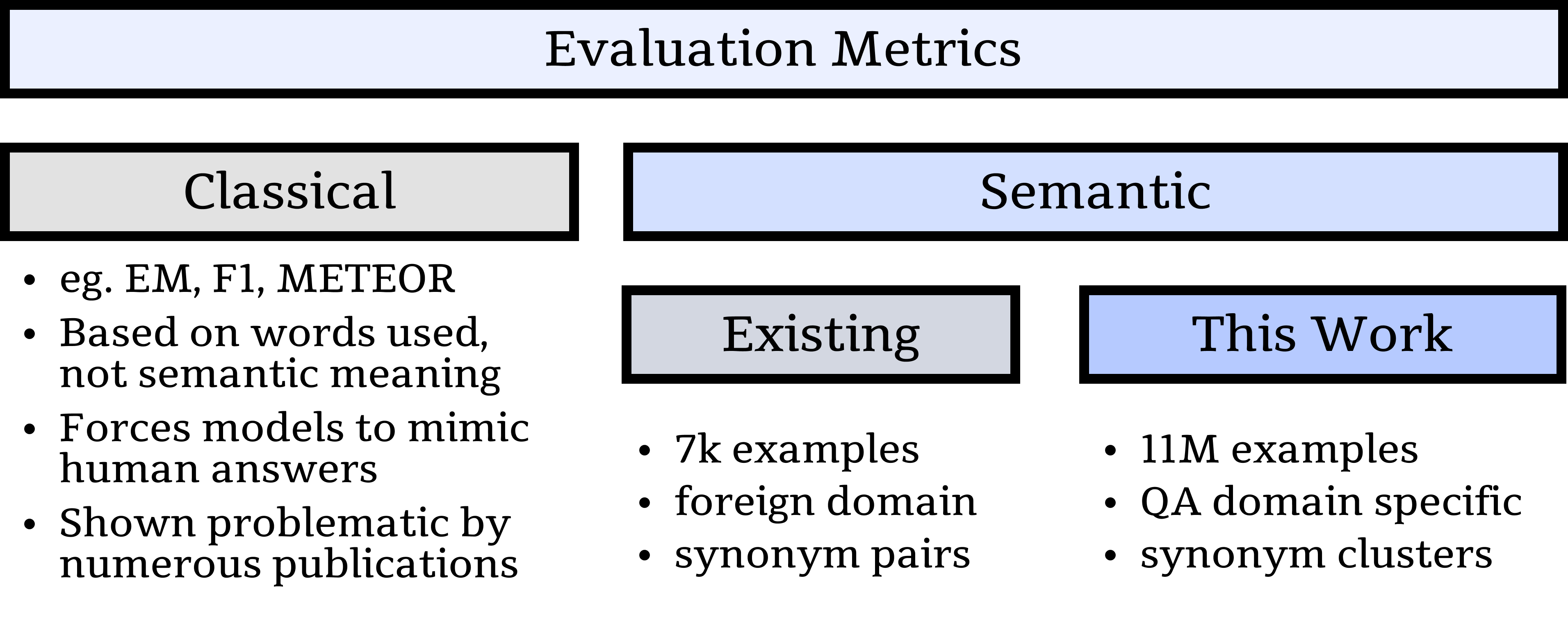}
  \caption{Graphical comparison of the most similar existing QA metrics.}
  \label{img:metric-comparison}
\end{figure}

\begin{figure*}
  \includegraphics[width=\textwidth]{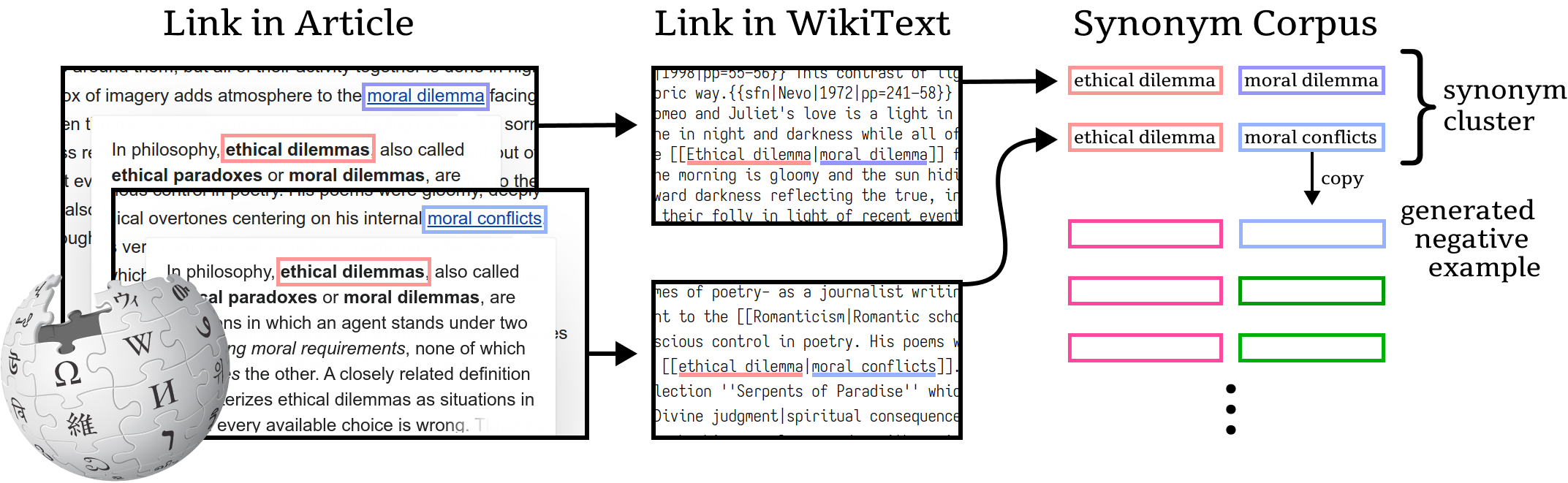}
  \caption{The WES generation pipeline. (article title, link text) pairs are extracted from the wikitext, then negative examples are generated by pairing article titles with link texts from other article titles. Groups of link text that are associated with the same article are all synonyms. These "synonym clusters" can be used to train evaluation metrics that consider multiple ground-truth answers simultaneously.}
  \label{img:generation-diagram}
\end{figure*}

In this paper, we introduce the Wiki Entity Similarity dataset: a large, task-specific semantic similarity dataset to train question-answering metrics. WES has over 11 million high-quality examples of synonymous and non-synonymous entities and phrases. In addition, WES is systematically mined from a democratic data corpus (Wikipedia), reducing bias from individual authors or annotators. Furthermore, WES generates synonym clusters of multiple phrases, which can be broken down pairwise into the classic semantic similarity formulation or grouped to simulate multiple ground-truth answers. Despite being auto-generated, human annotators consistently agree with WES labels. We show that a basic cross-encoder trained on WES is better at predicting human answers of correctness than classic metrics.

For the remainder of this paper, “outputs” are produced by QA models and “answers” are the ground-truth annotations that the model is trained on. In section 2, we take a closer look at current QA evaluation metrics and analyze their weaknesses. In section 3, we describe the generation of the novel WES dataset. In section 4, we characterize WES and present a baseline model trained on it. Finally, we conclude in section 5 with directions for future work.

\section{Related Work}

Although there are numerous desirable qualities for question answering models, we focus on evaluating the factual or conversational correctness of answers. Other metrics exist for testing information content, answer coverage, and other desirable qualities; however, this work is orthogonal to such metrics and should be used in conjunction with them. 
This section details existing metrics for evaluating the correctness of a model-generated output based on human-generated ground-truth answers. 


\textbf{Exact Match (EM)} is a standard binary metric for evaluating extractive QA. Before comparison, outputs and answers are often normalized by removing punctuation and stop words and converting to lowercase, as in \citep{SQuAD}. This metric relies on exact lexical matching: outputs that contain synonyms or informational prepositional phrases are considered incorrect by this metric. 

\textbf{F1} is a standard floating-point metric that measures the token overlap between the output and answer. It punishes longer, more informative outputs because such answers are less likely to exactly match the annotated ground truth. Finally, although F1 is more forgiving than EM, it fails to differentiate lexical and semantic dissimilarities. As a result, there is no threshold at which the F1 score correlates well with correctness. 

EM and F1 are the most prevalent question answering metrics today, serving as the leaderboard metrics for many of the most popular datasets (\citealp{SQuAD}, \citealp{NQ}, \citealp{HotpotQA}, \citealp{QuAC}). However, numerous works have found that lexical metrics can diverge from human judgements of model performance (\citealp{efficientqa}, \citealp{evaluating-qa-eval}, \citealp{SAS}).

\textbf{BLEU, ROUGE-N, ROUGE-L, METEOR}: \citep{BLEU}, \citep{ROUGEN}, \citep{ROUGEL}, \citep{METEOR} are common Natural Language Generation metrics based on measuring the n-gram overlap between output and answer. For instance, ROUGE-L measures the longest common subsequence and METEOR weights tokens by importance before applying an F1 analog. Like F1, these metrics punish longer, more detailed answers and unduly punish optional but informative prepositional phrases in model outputs. 

In summary, each of the above metrics relies on lexical matching to perform similarity checks. As a result, these metrics unfairly punish longer, more informative outputs that have unnecessary but useful prepositional phrases, and are particularly sensitive to the exact wording of the ground-truth annotation. Furthermore, the above metrics fail to model clause order and logical dependency in model outputs, which advantages illogical outputs that happen to contain words from the ground-truth answer. In recent years, newer metrics have been proposed that use contextual or semantic embeddings to compare sequences.  

\textbf{BERTScore} \citep{BERTscore} compares BERT contextual embeddings instead of raw tokens, which may more accurately capture underlying meanings and long-term dependencies. However, BERTScore still unfairly punishes optional but useful prepositional phrases as it is fundamentally a token-wise comparison. Finally, \citep{evaluating-qa-eval} has shown METEOR to correlate better with human judgment than BERTScore.
 
\textbf{Semantic Answer Similarity (SAS)} \citep{SAS} takes an end-to-end approach to evaluating generative QA. They compare bi-encoder, cross-encoder, and BERTScore approaches to QA evaluation, and find that a cross-encoder trained on answer similarity correlates most strongly with human judgment. However, SAS is trained on the Semantic Textual Similarity Benchmark \citep{stsb}, which has only 7k foreign-domain training examples. The creation of such semantic similarity datasets is tedious and susceptible to bias, and the small dataset size means models are prone to overfitting. We aim to improve on this cross-encoder approach by pretraining on a new, large, domain-specific, auto-generated dataset.

In summary, neural metrics are an emerging standard for more accurate question-answering evaluation. We aim to improve neural QA evaluation by training a new metric on a novel, large, QA-specific dataset that encodes entity similarity rather than general sentence similarity. 

\section{Wiki Entity Similarity Dataset}

The WES dataset is generated by pairing link text and target article text for links in Wikipedia. We generate our dataset in two stages: filtered link collection and negative example generation. 

\subsection{Link Filtering}

Following \citep{DPR}, we mine links from the English Wikipedia XML dump from Dec. 20, 2018. The XML format encodes all links in the form `[target article title|displayed link text]`, where the link will be displayed on the source page. We search directly for links that match the following criteria: \\
\textbf{Article title contains no hashtag} as hashtags denote links to subsections of Wikipedia articles.  \\
\textbf{Article title contains no colon} as colons denote pages under internal categories, such as User, User talk, Category, etc. \citep{WikipediaStyle} \\
\textbf{Article title contains no parentheses} as most (90\% of 500 randomly sampled) links to these articles simply remove the parenthetical instead of containing a synonym of the article title.

See Table \ref{tab:link-filtering} for examples of the motivation behind filtering out links with the above characters. 

\begin{table}
    \centering
    \begin{tabular}{cll}
    \hline
    \textbf{Rule} & \textbf{Sample Title} &\textbf{Sample Link Text} \\
    \hline
    \verb+#+ & \verb+VAX#Name+ & \verb+VAX trademark+\\
    \verb+:+ & \verb+WP:TWINKLE+ & \verb+TW+\\
    \verb+(+ & \verb+Styx (Band)+ & \verb+Styx+\\
    \hline
    \end{tabular}
    \caption{Motivation for excluding links to article titles that contain certain characters.}
    \label{tab:link-filtering}
    
\end{table}

After character-level filtering, we use additional heuristics to improve the quality of the link corpus.\\
\textbf{Deduplication}: we remove duplicate (article title, link text) pairs to improve dataset balance. Link texts with the same content but different capitalization are considered different to preserve the semantic meaning of capitalization in proper nouns. \\
\textbf{Incoming link threshold}: we enforce that each link in the corpus must link to an article that has a minimum of parameter n incoming links (citations) to filter out rarely used stub articles. Higher values of n ensure higher quality articles and a larger, more representative sample of synonyms for each article title. 
 \\
\textbf{Dictionary word linktext}: we remove links that have a dictionary word for the link text but a named entity or phrase for the target article title as these dictionary words are rarely fully qualified synonyms to the linked article title. Words are considered to be dictionary words if their WordNet \citep{wordnet} lemmatization is contained in the Python \verb+nltk+'s \citep{nltk} \verb+words+ corpus.

We experiment with link threshold values \(k=5\), \(k=10\), and \(k=20\), obtaining link corpora of 5,787,081, 4,407,409, and 3,195,545 distinct synonym pairs, respectively. 

\subsection{Negative Example Generation}
After harvesting synonyms from the Wikipedia links, we generate negative (non-synonymous) title-text pairs by matching article titles with link texts from other articles. The negative examples for an article title with \(k\) link synonyms are generated by sampling \(k\) other articles from the link corpus, then sampling one associated link text from each article to be paired with the article title (see how a negative example in the pink synonym cluster is created by copying a blue link text from a different article in Fig. \ref{img:generation-diagram}). This results in each article title being associated with the same number of non-synonyms as distinct synonyms, preserving dataset balance. Choosing non-synonyms from different articles ensures a representative covering of the sample space and precludes duplicate negative examples. After negative example generation, the largest version of WES (with incoming link threshold \(k=5\)) has 11M+ examples, and the highest quality version (with \(k=20\)) contains 6.4M examples. 

Future iterations of the WES dataset may be made more difficult through adversarial generation. Such methods condition negative example generation on the article title to create negative examples which are more similar to but not synonymous with the paired phrase. Possible future generation techniques include:\\
\textbf{F1 ranking}: negative examples generated from the most lexically similar alternatives will create textually similar but semantically different examples, increasing difficulty for token-based models. \\
\textbf{Substring selection}: negative examples generated from short substrings of the article title may be synonymous in context, but will not be specific enough to globally qualify a concept. Such examples force models to watch for appropriate specificity. \\
\textbf{Word co-occurrence}: negative examples generated from non-synonymous words with high co-occurrence will create pairs of grammatically feasible phrases, increasing difficulty for part-of-speech-based models. \\
\textbf{Article co-occurrence}: the titles of often co-current but semantically distinct articles can be treated as very difficult non-synonymous examples, teaching models to differentiate between parts of speech and improving sensitivity.

\section{Analysis}

We use human evaluation to check the quality of the collected WES dataset. For each task, we randomly sample 25 positive and 25 negative training pairs from the link threshold \(k=20\) dataset. Two annotators rate the synonymy of the pairs following the ranking scheme used in \citep{SAS}. We treat annotated scores of 2 or 3 as “synonymous” and 1 as “non-synonymous,” and find that only 2\% of labels are incorrect. Dataset label accuracies and inter-annotator agreement from different stages in the filtering process are listed in Table 1.

\begin{table}
    \centering
    \begin{tabular}{lcc}
    \hline
    \textbf{Dataset} (link threshold $k = 20$) & \textbf{Acc} & \textbf{$r$} \\
    \hline
    Full Dataset & 98\% & 0.932\\
    Without Dictionary Filtering & 91\% & 0.908\\
    \hline
    \end{tabular}
    \caption{Average accuracy and Pearson's $r$ correlation between annotators in the full dataset and ablations.}
    \label{tab:human-eval}
    
\end{table}
\section{Conclusion}
We introduce WES, an 11M example semantic entity similarity dataset for training question answering evaluation models. WES is generated by treating Wikipedia link texts and target article titles as synonyms then filtering for quality. WES is targeted to question answering evaluation, independent of human annotators, and consistent with human judgment.
We hope that future question-answering datasets will implement semantic evaluation metrics in their leaderboards to encourage the development of more free-form models. In future works, link-mining similarity datasets like WES can be made more challenging by generating negative examples adversarially as described at the end of section 3, more consistent by unioning semantic clusters according to Wikipedia’s internal redirect pages, and more comprehensive by leveraging link-to-link pairwise synonymy within semantic clusters. 

\section*{Acknowledgements}

We would like to sincerely thank Di Jin for his guidance during the literature review and method design process, Lufan Wang and Houjun Liu for their feedback on data collection methods and help proofreading this paper, and Yan Liu, Qiong Huang for their help with dataset analysis and compute. We would also like to thank \citep{si2021} for reminding us to have some fun in our literature. 
\bibliography{anthology,custom}
\bibliographystyle{acl_natbib}

%
%

\end{document}